# A Hybrid Data-Driven Approach For Analyzing And Predicting Inpatient Length Of Stay In Health Centre

Tasfia Noor Chowdhury [a], Sanjida Afrin Mou[a], Kazi Naimur Rahman[a,*]

[a] Department of Mechatronics & Industrial Engineering, Chittagong University of Engineering & Technology (CUET), Chattogram 4349, Bangladesh

A B S T R A C T

Patient length of stay (LoS) is a critical metric for evaluating the efficacy of hospital management. The primary objectives encompass to improve efficiency and reduce costs while enhancing patient outcomes and hospital capacity within the patient journey. By seamlessly merging data-driven techniques with simulation methodologies, the study proposes an all-encompassing framework for the optimization of patient flow. Using a comprehensive dataset of 2.3 million de-identified patient records, we analyzed demographics, diagnoses, treatments, services, costs, and charges with machine learning models (Decision Tree, Logistic Regression, Random Forest, Adaboost, LightGBM) and Python tools (Spark, AWS clusters, dimensionality reduction). Our model predicts patient length of stay (LoS) upon admission using supervised learning algorithms. This hybrid approach enables the identification of key factors influencing LoS, offering a robust framework for hospitals to streamline patient flow and resource utilization. The research focuses on patient flow corroborates the efficacy of the approach, illustrating decreased patient length of stay within a real healthcare environment. The findings underscore the potential of hybrid data-driven models in transforming hospital management practices. This innovative methodology provides generally flexible decision-making, training, and patient flow enhancement; such a system could have huge implications for healthcare administration and overall satisfaction with healthcare.

## 1. Introduction

Health systems worldwide face such scarcities as excess patient admission, lack of resources in public hospitals, and high costs with private healthcare. Length of Stay (LoS) is one of the key metrics in gauging hospital efficiency, and it affects operations, availability of beds, and even patient outcomes. Accurate prediction of LoS can foster optimum resource utilization, reduction of delays, and enhance patient satisfaction.

Cheng and Hu (2009) [1] underscored the impact of LoS on care costs and service quality, forming the basis for predictive systems. Fang et al. (2022) [2] proposed a hybrid model combining convolutional neural networks and support vector machines for predicting traumatic brain injury outcomes. He et al. (2019) [3] highlighted simulation modeling's dominance in bed management, emphasizing interunit interactions during peak volumes. Ortiz-Barrios et al. (2024) [4] examined AI and simulation to mitigate bed shortages, identifying Random Forest as an effective predictive tool.

Chiara Cirrone [5] utilized Lean Six Sigma and super network frameworks to optimize emergency department flow. Kieran Stone [6] emphasized comprehensive frameworks for LoS prediction using diverse parameters, while Deepak Yaduvanshi [7] tackled outpatient queueing issues via queueing theory and SWOT analysis. Studies by Bean et al. [8] and Crilly et al. [9] highlighted tools like the Patient Admission Prediction Tool (PAPT) for improved patient flow, although limitations like localized data and communication gaps remain. Lei Zhao [10] applied queueing theory and Markov Chains, and Xin Ma and Yabin Si [11] utilized Just-in-Time Learning with one-class Extreme Learning Machine for LoS prediction.

Chuang et al. [12] demonstrated machine learning classifiers, including Random Forest, SVM, and Gradient Boosting, for LoS predictions. Baek et al. (2018) [13] analyzed clinical and administrative variables affecting LoS, noting that diagnoses like cerebral infarction and myocardial infarction prolonged stays. Breda et al. (2022) [14] evaluated the Geriatric Fracture Program, demonstrating reduced LoS and costs. Bhadouria and Singh (2023) [15] integrated patient profiles and machine learning for LoS and mortality prediction, enhancing decision-making.

Longo et al. (2024) [16] reviewed AI/ML applications in orthopedic LoS predictions, while Guo et al. (2023) [17] explored the impact of social determinants of health on LoS during COVID-19, emphasizing comprehensive data sharing. Sivarajkumar et al. (2023) [18] addressed bias in patient data models, enhancing equity in predictions. Young et al. (2023) [19] identified risk factors in SARS-CoV-2-positive pregnancies using Random Forest, achieving high accuracy. Grill et al. (2011) [20] linked goal attainment in geriatric rehabilitation to improved functioning and shorter LoS.

Ismail et al. (2012) [21] analyzed predictors of LoS in mood disorder patients, finding factors like living alone and frequent admissions to prolong stays. Roshanghalb et al. (2024) [22] proposed a data-fusion model integrating diverse clinical data, achieving superior performance. Chen et al. (2024) [23] employed deep learning techniques, combining convolutional neural networks and long short-term memory models for accurate LoS predictions. Lastly, Chen et al. (2023) [24] addressed patient data heterogeneity with a robust data-fusion model.

Our research study proposed a hybrid data-driven approach integrating machine learning, process mining, and simulation techniques for LoS predictions. Key determinants will be identified through the analysis of various datasets and scalable solutions will be proposed. Results from this study would aid the hospitals to predict patient needs, optimize resources, and enhance operational efficiency.

---

*Abbreviations:* LoS, Length of Stay; PLoS, Prolonged Length of Stay; PAPT, Patient Admission Prediction Tool; EHRs, Electronic Health Records; RTDC, Real-Time Discharge Capacity; GBM, Gradient Boosting Machine; PCA, Principal Component Analysis; EFB, Exclusive Feature Bundling; APR, All Patient Refined; CCS, Clinical Classifications Software; DRG, Diagnosis Related Groups; MDC, Major Diagnostic Category; LoR, Logistic Regression; SVM, Support Vector Machine.

\* Corresponding author.
   *E-mail addresses:* u1809001@student.cuet.ac.bd (T.N. Chowdhury), 1809025@student.cuet.ac.bd (S.A Mou), naimur@cuet.ac.bd (K.N. Rahman)

## 2. Method

Length of stay (LoS) in the hospital measures the number of days an inpatient spends in the hospital and is critical for improving patient care, reducing costs, and optimizing resource utilization. LoS differs widely among patients with the same condition, depending on patient characteristics, complications, and treatment complexity. Reliable LoS prediction allows for better healthcare planning such as staff allocations, admissions, and optimizing the use of beds while improving operational efficiency. Reliable models will enable hospitals to adopt preventive measures that may reduce the risk of falls, infections, and medication errors that usually occur during prolonged stays. For patients, shorter LoS promotes quicker recovery in preferred settings like home care. From a healthcare provider's perspective, reducing LoS improves resource allocation, benefiting more patients. While prior studies categorize patients by health issues with suggested LoS ranges, accurate prediction remains challenging due to diverse influencing factors. This paper proposes a robust LoS prediction approach to ensure consistent, actionable insights across healthcare settings, with **Fig. 3** presenting the study's workflow.

2.1 LoS & PLoS

Length of Stay (LoS) refers to the stay of a patient in the hospital from admission to discharge. It is a crucial measure most hospitals utilize for understanding and managing patient flow, resource allocation, and overall efficiency within the hospital. LoS has commonly been used to evaluate the quality of care, improve hospital operations, and manage the costs of healthcare. Prolonged Length of Stay (Prolonged LoS) occurs when a patient's hospital stay extends beyond the expected or standard duration for their specific diagnosis or procedure. This extended stay can result from various factors such as complications, comorbidities, delays in treatment, discharge planning issues, or social circumstances. Prolonged LoS can indicate potential inefficiencies in care delivery and may lead to increased healthcare costs, higher risk of hospital-acquired infections, and greater strain on hospital resources. Reducing prolonged LoS is a key objective for healthcare providers to improve patient outcomes and optimize resource utilization.

2.2 Patient Flow

**Fig.1.** presents the Patient flow network topology to the structure and interconnectedness of various units, departments, and pathways within a healthcare facility that patients navigate during their care journey. Topology may include different units such as emergency departments, outpatient clinics, inpatient wards, operating rooms, diagnostic centers, and other specialized areas. It also encompasses the pathways and transition patients follow, from admission and triage to treatment, monitoring, and eventual discharge or transfer.

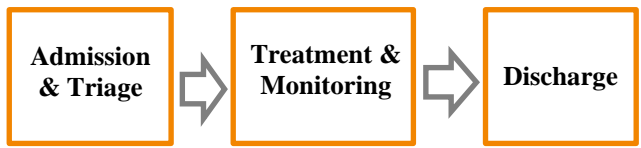

Fig. 1. Patient Flow Network

2.3 Cause & Effect - Ishikawa Diagram

The central problem "Long Queued Patient Flow in Public Hospital" is shown in the diagram. The main causes are categorized into four groups: Process Factors, Resource Factors, Communication Factors, and External Factors (shown in **Fig.2**.)

Each cause is further divided into sub causes that contribute to the extended waiting times for patients in a public hospital. Some causes are:
- An inefficient triage process (cause) can lead to prolonged waiting times for patients (effect).
- Insufficient staffing in inpatient wards can lead to overcrowding and delays in allocating beds for admitted patients.
- Poor communication between different departments responsible for patient care transitions.

- If scheduling of diagnostic procedures and tests is done indiscriminately without coordination, it would lead to a longer stay of patients waiting for test results. This could affect the flow of patients and increase resource utilization.

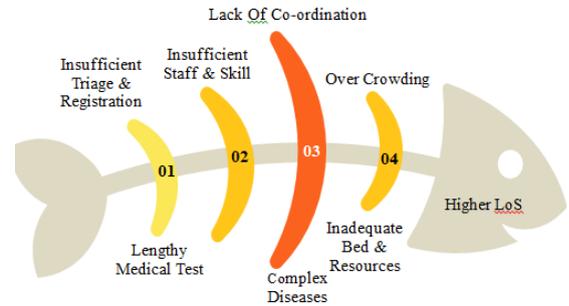

Fig. 2. Cause & Effect diagram for Patient flow in hospital system

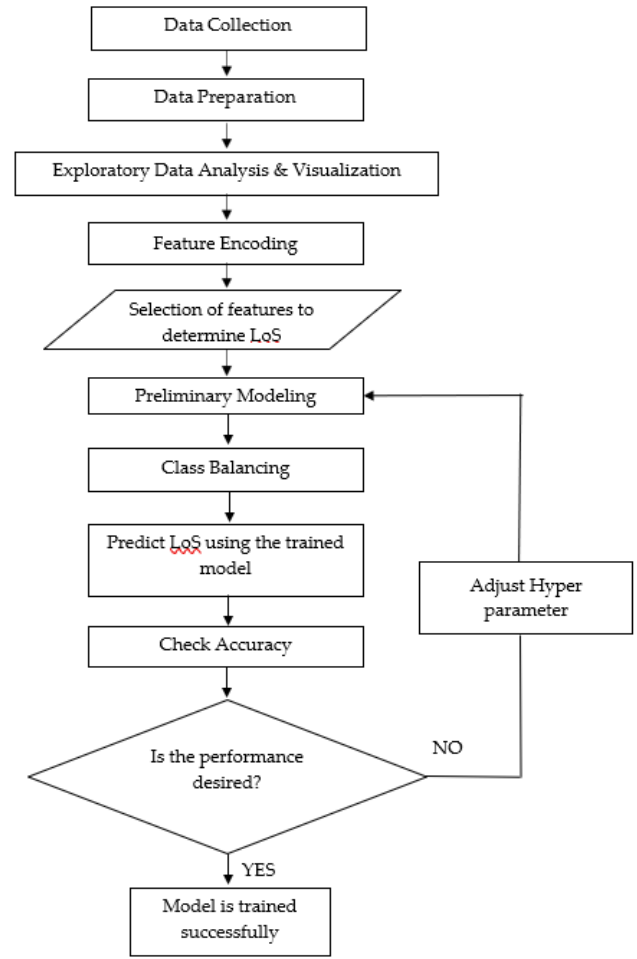

Fig. 3. Prediction of LoS Workflow Diagram

2.4 Data Collection

A wide variety of data types are needed in order to anticipate patient's LoS using machine learning. The dataset, available at New York State Government Health Data Website and provided by NY Department of Health includes over 2.3 million rows of de-identified inpatient discharge information. This data contains a variety of variables associated with individual patient information, medical codes, and hospital details, including Service Area, Facility ID, Age Group, Gender, Length of Stay, Type of Admission, and diagnostic and procedural codes. Such abundant data lend themselves to numerous investigations, including machine learning models predicting hospital LOS. This rich data is essential for designing accurate predictive models that will support health-care planning and resource allocation.

2.5 Data Pre-Processing and Visualization

Before any analysis starts, exploring the dataset is instrumental in making better-informed decisions. It also allows the determination of issues such as missing values or outliers. At the same time, it assists in the selection of the right statistical techniques, helping people observe trends and relationships.

Process of data cleaning: Data cleaning is the exercise of cleaning information such that issues include some aspects of missing values, outliers in a dataset, and noise within it. This process may include techniques like imputing missing values, removing outlier values, or filtering out noisy data points to ensure high levels of quality or reliable dataset. We removed the variables of License Number, Payment Typology 2,3, Zip Code - 3 digits, Facility Id, Attending Provider License Number, which did not aid our analysis. Further, we examined the relationship between the numerical variables and established those by correlation heatmaps.

From our first glance at the correlation map, it's clear that some features are tightly connected, and importantly, with the length of stay. While not all columns in the correlation matrix are meaningful to analyze we can spot some notable relationships. The severity of illness code, along with total costs, shows a strong positive correlation with the LoS.

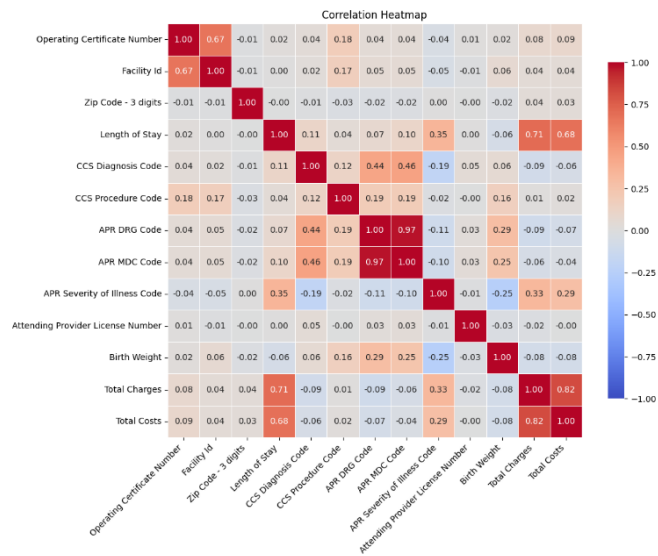

Fig.4. Correlation Heatmap

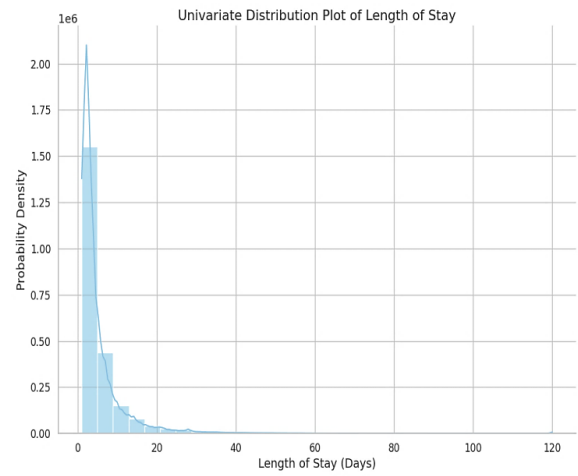

*Fig.5. Univariate Distribution Plot of LoS*

The **Fig. 5** depicts LoS values range from 1 to over 120 days, with stays of 120 days or more aggregated into a single category. The distribution is highly skewed, as the majority of patients have stays between 1 and 5 days.

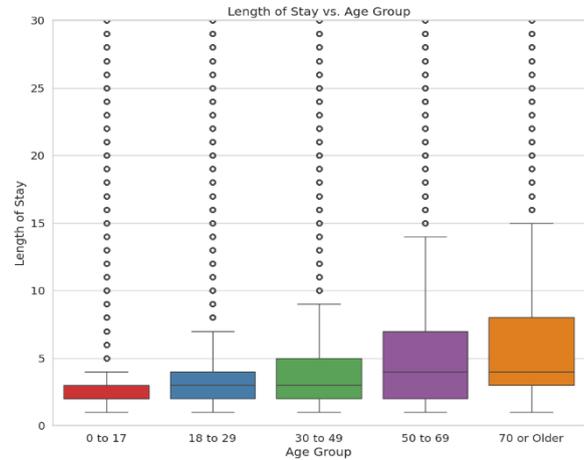

Fig.1.  Age Group Vs Length of Stay

**Fig.6.** shows that the tendency of longer stay as the age increases gradually.

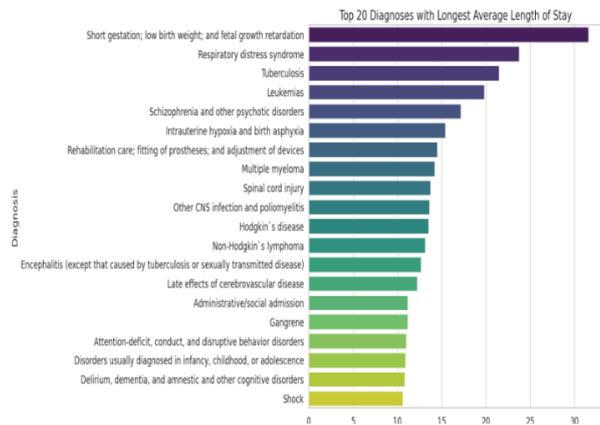

Fig.7.  Top 20 Diagnoses with Longest Average Length of Stay

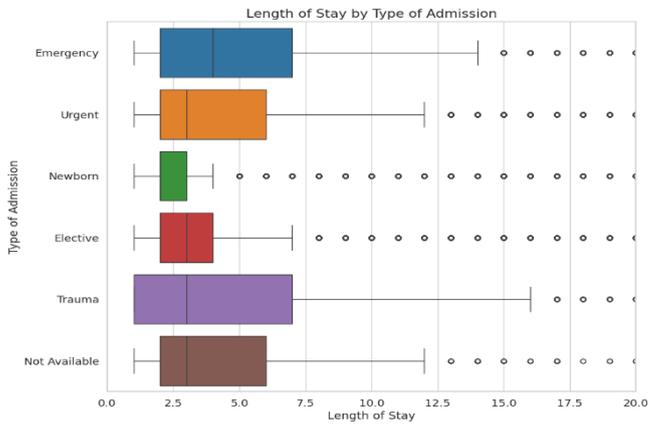

Fig.8. Length of Stay Vs Type of Admission

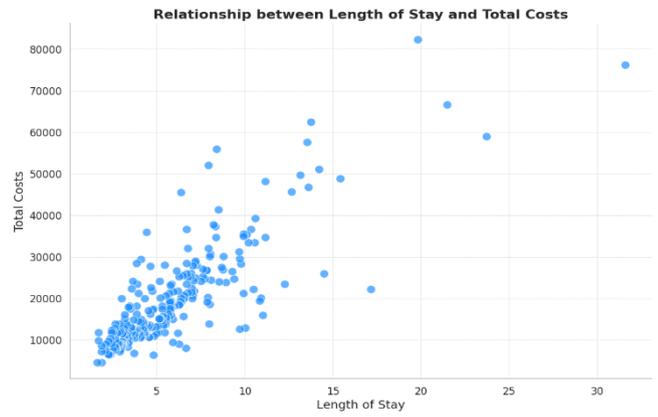

Fig.10. Scatter Plot of LoS Vs Total Cost

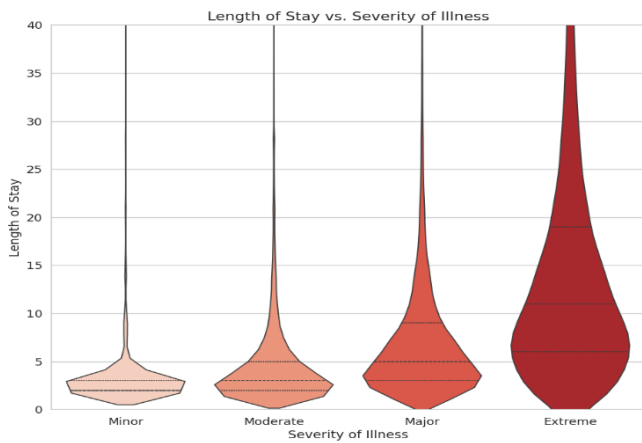

Fig.9. Violin Plot of Severity of Illness vs Length of Stay

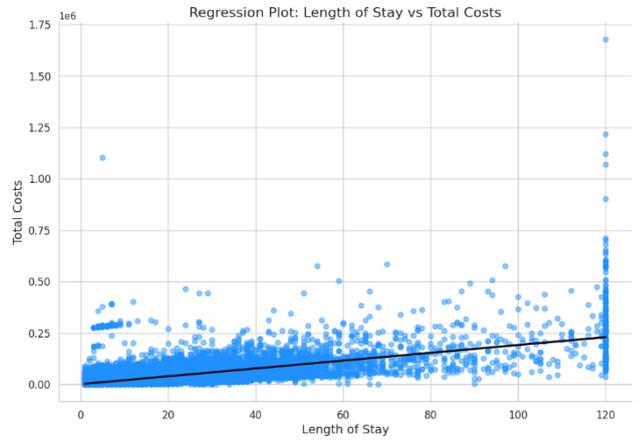

Fig.11. Length of Stay Vs Total Cost

**Fig.8**. presents a significant degree of variation is prominent within the Severity of Illness category, suggesting that it would probably be a crucial component of machine learning (ML) models.

Performed an inner join (index), the relationship between diagnosis descriptions that incur the highest costs, and their corresponding lengths of stay can be examined. This comparison reveals a strong relationship between the most expensive diagnoses and longer inpatient length of stay. However, there are exceptions. For instance, the diagnosis with the second longest average length of stay ranks only as the fifth most expensive to the hospital. This suggests that while cost and length of stay are generally correlated, other factors also influence hospital expenses.

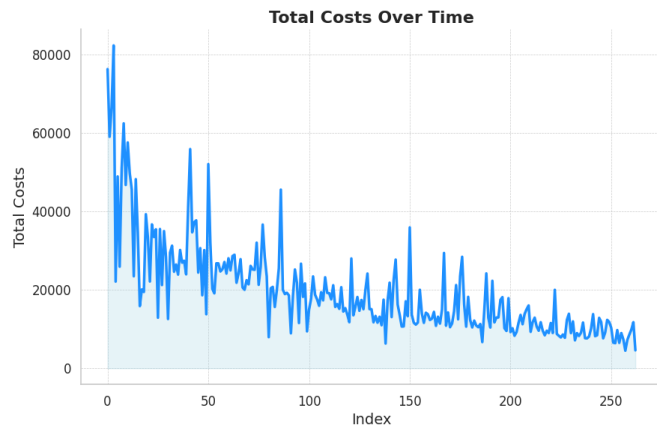

Fig.12. Relationship between Index & Total Cost

**Fig.2**. shows the positive linear relationship between LoS and costs associated. The slope of this correlation shows that as the stay increases, costs also rise.

2.6 Feature Encoding

Feature encoding involved transforming categorical columns containing strings. Categories with corresponding codes, like Severity of Illness, were numerically encoded. String indexing was used for a selection of characteristics, including Age Group and Risk of Mortality. For instance, '0 to 17' received a code of 1, indicating a younger age group. Similarly, 'Minor' was encoded as 1, while 'Extreme' received 4 in APR Risk of Mortality. String indexing helps maintain dataset dimensionality and lets models learn the order among feature categories.

2.7 Applying Bins

We use a multi-class classification strategy to predict hospital stays lasting from one to 120 days. Rather than treating each day as a separate class, we divide them into relevant categories. This strategy strikes a balance between model accuracy and forecast specificity. Our final bins are as follows: 1-5, 6-10, 11-20, 21-30, 31-50, and 50-120 days. The Fig: 13 clearly indicates a huge class imbalance in the dataset.

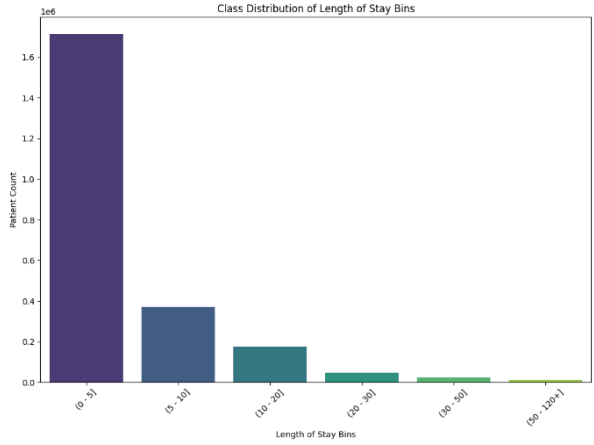

Fig.13. Patient Counts vs Length of Stay Bins

2.8 Principal Component Analysis

Principal Component Analysis (PCA) is a versatile technique that can handle datasets with various characteristics like correlated variables, missing data, and outliers. Its goal is to identify the key information in the data and express it as principal components - summary indices that capture the most variance. PCA finds the principal part that are orthogonal linear combinations of the actual variables, with the first component capturing the most variance.

Here, PCA was applied after normalizing the train and test data with StandardScaler(). It is useful for reducing dataset dimensionality, which is beneficial for large datasets, while also removing multicollinearity. Although not primarily designed for mixed numerical/categorical data, PCA's performance wasn't significantly impacted here. Importantly, data should always be normalized before PCA because it projects raw data onto variance-maximizing directions by calculating relative distances within features. When features have different scales, PCA can be skewed.

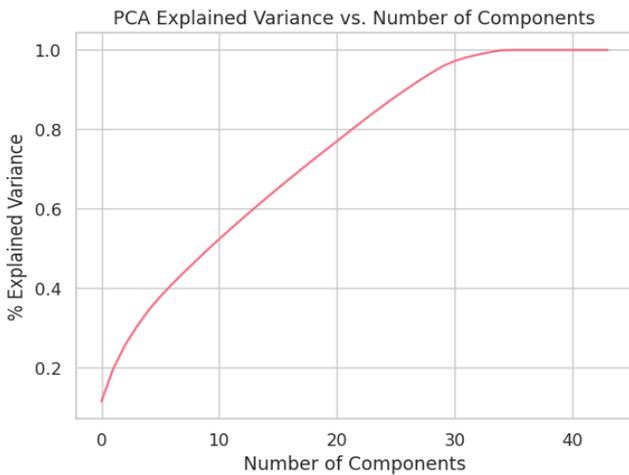

Fig.14. Variance Vs No. of Components

To decide the number of main components to keep, the share of variance explained was plotted versus the number of factors from PCA. From this plot, 29 components were selected as the minimum required to cover 95% of the variation, allowing for dimensionality reduction while retaining most of the critical information and avoiding overfitting or noise from too many features.

2.9 Feature Selection

Feature selection is a method that aims to improve model performance by identifying and retaining only the most related data features while discarding unnecessary or redundant ones. It involves automatically selecting the input variables for the machine learning model that are most important. From the feature importance plot, we can see that certain feature, such as the age group, type of admission, risk of mortality, payment typology, and emergency department indicator, play a significant role in explaining the variance in the dataset. These features are likely capturing key information related to the target variable we are trying to predict, and therefore, should be given higher priority during model building and interpretation.

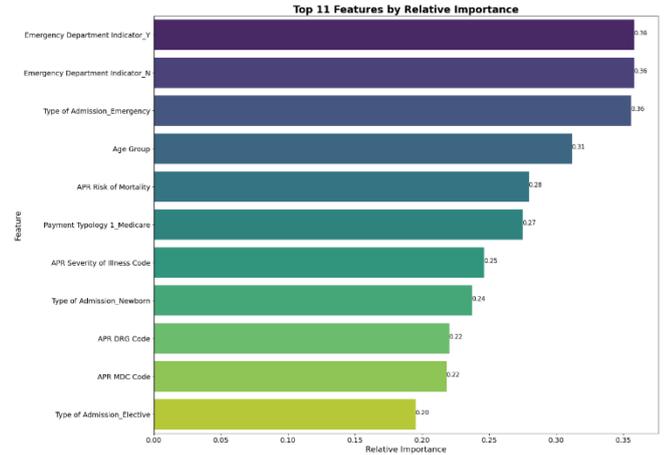

Fig. 15. Top 11 Feature Selected by Relative Importance

2.10 ML Models Used

In our research, we sought to compare different models of machine learning to predict stay length in hospital; accordingly, each highlighted its own positives and unique contribution.

Logistic Regression (LoR) is a simple yet powerful model that is often used in binary classification tasks. It estimates the probability of an outcome using maximum likelihood estimation. The simplicity and easy-to-interpret nature of the LoR make it a great tool for interpreting which of the different input features have an impact on predictions.

Decision Trees (DTs) provide an intuitive and visual method of addressing classification and regression problems. They accomplish this using a "divide-and-conquer" technique whereby information is progressively broken down into its smaller constituents. It starts with a root node by which the entire data set is covered, each decision node being linked to a feature-based question, and each leaf node is then the final classification prediction. The idea is to split the input data such that different classes will not mix, using metrics such as the GINI index and Information Gain to work out optimal splits A few hyperparameters can be varied to fine-tune the performance and efficiency of the Decision Tree. In particular, the general attributes are: n_estimators-hence the number of trees built before averaging predictions; max_features-hence the number of features that are maintained while calculating the split; min_samples_leaf-minimum samples necessary for splitting an internal node; max_leaf_nodes-maximum number of leaf nodes allowed per tree. The adjustment of these hyperparameters allows tuning the model's behavior and thus improving its efficiency.

Entropy of a node S, $H(S) = -P_{(0)} \log_2 P_{(0)} - P_{(1)} \log_2 P_{(1)}$ (1)

$$P_{(1)} = \frac{P_{(1)}}{P_{(1)} + P_{(0)}}$$ (2)

Gini index for node S = $1 - \{(P_1)^2 + (P_0)^2\}$ (3)

Information Gain, $IG(D, f) = H(D) - \frac{N_{left}}{N} H(D_{left}) - \frac{N_{right}}{N} H(D_{right})$ (4)

Random Forests improve on decision trees by building multiple trees and averaging their predictions. This strategy decreases the risk of overfitting, which occurs when a model is trained well but does badly on previously unseen data. By averaging the results of many trees, Random Forests provide more accurate and stable predictions.

Final Prediction, $\hat{y} = \text{mode } \{T_1(x), T_2(x), T_n(x)\}$ (5)

AdaBoost enhances weak learners by putting more emphasis on cases previously misclassified. The weights of the cases are adjusted in each iteration to improve accuracy. Such an approach leads AdaBoost to combine weak models to yield a strong predictive model and thus enhance performance.

Gradient Boosting builds models sequentially to minimize the prediction errors. Using decision trees as base classifiers, it refines the model sequentially by addressing the residual errors of earlier models to improve accuracy.

Final Prediction, $\hat{y}(x) = \hat{y}_0 + \sum_{m=1}^{M} \eta h_m(x)$ (6)

LightGBM enhances training and reduces memory utilization via novel approaches such as GOSS and EFB.

Objective Function: $Obj(t)^j = \sum_i^n \left[ g_i f_j(x_i) + \frac{1}{2} h_i f_j^2(x_i) \right] + \Omega(f_j)$ (7)

## 3. Result

In this section we have evaluated the following ML models namely Logistic Regression, Decision Tree, Random Forest, AdaBoost, and LightGBM to predict Hospital Length of Stay treated as a multiclass classification problem. With the definition of meaningful class bins together with robust evaluation metrics, such as Cohen's Kappa, MCC, F1-score, recall, accuracy, and precision, we guaranteed correct and interpretable predictions. Our methodology offers a full-scope focus on predictive modeling in healthcare, emphasizing model selection, tuning, and robust evaluation, which ensured the generation of reliable and actionable predictive insights leading to informed decision-making in a clinical setting.

LoS is categorized into discrete bins (e.g 1–5, 6–10) so we use classification metrics.

Accuracy: The percentage of accurately classified incidents.

$$\text{Accuracy} = \frac{\text{Number of Correct Prediction}}{\text{Total Number of Prediction}}$$

Precision: The ratio of accurately predicted positive findings to total anticipated positives.

$$\text{Precision} = \frac{TP}{TP + FP}$$

Recall: The percentage of accurately predicted positive observations to total observations in the actual class.

$$\text{Recall} = \frac{TP}{TP + FN}$$

F1-Score: The harmonic mean of Precision and Recall.

$$F1 - \text{Score} = 2 \frac{\text{Precision} + \text{Recall}}{\text{Precision} \cdot \text{Recall}}$$

Cohen's Kappa: A metric that compares observed accuracy to expected accuracy (by chance).

$$\kappa = \frac{p_0 - p_e}{1 - p_e}$$

Matthews Correlation Coefficient (MCC): A metric for the accuracy of multiclass classifications, that considers true and false positives and negatives.

$$\text{MCC} = \frac{TP \cdot TN - FP \cdot FN}{\sqrt{(TP + FP)(TP + FN)(TN + FP)(TN + FN)}}$$

Comparative assessments were conducted to discern the models' performance and generalizability.

3.1 Performance Comparison of ML Models

Table 1.

The prediction performance of the ML models for LoS prediction

| Methods | Logistic Regression | Decision Tree | Random Forest | Ada Boost | LightGBM |
|---|---|---|---|---|---|
| **Accuracy** | 0.73 | 0.61 | 0.65 | 0.76 | 0.78 |
| **Precision** | 0.76 | 0.93 | 0.92 | 0.84 | 0.89 |
| **Recall** | 0.98 | 0.70 | 0.74 | 0.80 | 0.84 |
| **F1 Score** | 0.86 | 0.80 | 0.82 | 0.82 | 0.83 |

Logistic Regression is good at finding the true positives (high recall) but has moderate accuracy and precision. The Decision Tree is very precise (few false positives) but misses more true positives (lower recall) and has lower overall accuracy. Random Forest improves recall compared to the Decision Tree but still has lower accuracy than Logistic Regression and LightGBM. AdaBoost is quite balanced with good accuracy, precision, recall, and F1 score. It performs better than LoR in terms of precision and slightly better in accuracy. LightGBM has the highest accuracy and precision and also strong recall and F1 score, making it the best overall performer.

LightGBM is the best model for predicting hospital length of stay. It has the highest accuracy (0.78) and precision (0.89), and strong recall (0.84) and F1 score (0.83). This means it makes the most reliable and accurate predictions overall. While Logistic Regression is very good at finding true positives, its overall accuracy and precision are lower than LightGBM. Decision Tree and Random Forest have high precision but lower accuracy and recall. AdaBoost is balanced but doesn't outperform LightGBM in any metric.

There are two further evaluation metrics: Cohen's Kappa and MCC. Kappa assesses the relation between predicted and actual classifications with chances. Its values vary from -1 (total disagreement) to 1 (perfect agreement), with 0 indicating no agreement other than chance.
The Matthews Correlation Coefficient (MCC) is a correlation coefficient that measures the relationship between observed and predicted binary classifications. It includes true positives, true negatives, false positives, and false negatives, resulting in a balanced score even when class numbers vary greatly.

Table 2.

ML Models Cohen's Kappa & MCC Score

| ML Model | Cohen's Kappa | MCC |
|---|---|---|
| **Logistic Regression** | 0.51 | 0.5 |
| **Decision Tree** | 0.44 | 0.49 |
| **Random Forest** | 0.69 | 0.56 |
| **Ada Boost** | 0.58 | 0.61 |
| **LightGBM** | 0.61 | 0.63 |

Among the models tested, Random Forest and LightGBM had the greatest Cohen's Kappa and MCC values, indicating that they perform best in terms of agreement and correlation with the real data. Ada Boost also performs well, although Logistic Regression and Decision Tree models perform only moderately. The choice of model may be determined by the task's specific criteria, such as the relevance of interpretability, resources, and necessity to deal with imbalanced data.

Table 3.

Hyper Parameter Tuning Summary

| ML Model | Hyper Parameter Tuning Description |
|---|---|
| **Logistic Regression** | C: 0.1, penalty: l2, solver: liblinear |
| **Decision Tree** | max_depth: 20, min_samples_split: 10, min_samples_leaf: 5, max_features: auto, criterion: entropy |
| **Random Forest** | n_estimators: 200 max_depth: 30, min_samples_split: 2, min_samples_leaf': 1, max_features: sqrt, bootstrap: True, random_ state = 42 |
| **Ada Boost** | n_estimators: 100, learning_rate: .1, base_estimator_max_depth: 3 |
| **LightGBM** | n_estimators: 300, learning_rate: 0.01, max_depth: 10, num_leaves: 63, min_data_in_leaf: 20 |

Table 3 shows the values of hyperparameter chosen by RandSearchCv for various models, demonstrating its ability to find effective parameter settings. Thus, the models are trained.

## 4. Discussion

This study presents a hybrid data-driven approach to predicting Length of Stay (LoS) in hospitals by integrating machine learning models and process mining techniques. The findings demonstrate that this methodology provides actionable insights into patient flow, resource allocation, and operational efficiency in healthcare management.

4.1 Key Findings and Model Performance

Among the models tested, (**Table 1**) LightGBM achieved the highest accuracy (78%) and precision (89%), making it the most robust and reliable predictor for LoS. This result supports the notion that advanced ensemble models like LightGBM, which incorporate gradient boosting, can handle complex healthcare datasets effectively. Comparatively, Logistic Regression exhibited strong recall but lacked the overall accuracy of LightGBM, while Random Forest and AdaBoost delivered balanced performance but fell short in metrics such as F1 score and precision.

4.2 Strengths of the Approach

The hybrid methodology is superior in various ways. Firstly, it efficiently functions on high-dimensional data using the Principal Component Analysis (PCA) for dimensionality reduction. This circumvents a lot of complexity in computation while retaining valuable information. Secondly, its inclusion of diversified features—essentially from APR severity codes to admission types—guarantees that it considers all factors affecting LoS.

4.3 Data-Driven Insights

The rich dataset that incorporated demographic, clinical, and procedural variables allowed the study to identify some of the main predictors of length of stay (LoS), such as:

- Age Group: Older patients had longer LOS, commensurate with their generally greater need for care.
- Severity: Patients having higher severity codes had longer lengths of stay, which reflects the relationship between clinical complexity and LoS.
- Type of Admission: Emergency admissions are associated with prolonged length of stay because of their unpredictable nature and complexity.

Although revealing the significance of the ML application to optimize healthcare organizations, it urges modernity to realize their widest possible applicability and equity in healthcare delivery. The future study and implementation of such methods may offer a distinct opportunity for change in managing healthcare for the better for consultants and the patients, too.

## 5. Conclusion

One major reason to reduce in length of stay in health care is to optimize resource use and improve patient outcomes. Our study revealed that, among the models applied for length-of-stay prediction, LightGBM was the most effective in forecasting. Thus, one has to keep in mind the proper training using the most suitable ML method for any given dataset and relevant problem. These measures would justify patent clinical early-stage decision-making solely on the performance factors of the algorithms for identifying optimal models. Within this study, both Gradient Boosting (GB) and Logistic Regression (LoR) has shown themselves to be the most effective in identifying a high-risk patient group that has the potential benefit from timely interventions that lead to reduced length of stay. It offers insight and the opportunity for informed decision-making for current and future patient care. However, by nature, there is no universal agreement on how one would decide the appropriate model for every specific context because different model choices depend on data type and analysis goals. A thorough evaluation and comparison are, however, to ensure the general applicability of models.

The study demonstrates the importance of the APR DRG Code as a vital predictor of LoS, which classifies patients based on the reason for their admission, illness severity, and risk of death. We built five action-based machine learning models for predicting LoS, then evaluated the output using performance metrics such as accuracy, precision, recall, and F1 score. LightGBM model was revealed to perform robustly and would be a suitable candidate in classification-related tasks in predictive analytics and healthcare decision-making. With these advantages, predictive models can, further, streamline healthcare providers in resource allocation, discharge planning, and patient management. Also, finally, it has been shown that there is 70% accuracy in the prediction of patient LoS based on who was admitted into the study, and this is an insight that could modify healthcare management.

Future research should design sequential models to estimate total treatment costs based on LoS to help resource allocation and financial planning for health care systems. Advanced predictive tasks such as LoS prediction after surgeries and critical hospital resources consumption optimization should be covered as well. Additionally, development could be made on a robust management system to interface various data analytics to support real-time planning and decisions for health care centers. These enhancements will enable improved efficiencies in the delivery of health care and patient care.

**Statements of ethical approval**

This article does not contain any studies with human participants or animals performs by any of the authors.

**Declaration of Competing Interest**
The authors declare that they have no conflict of interest.

**Supplementary material**
Supplementary material associated with this article can be found in https://shorturl.at/k9stt